\documentclass[10pt, a4paper]{article}

\usepackage[dvipsnames]{xcolor}

\usepackage{lrec-coling2024} 

\usepackage{comment}
\usepackage{paralist}
\usepackage{url}
\usepackage{booktabs}
\usepackage{graphicx}
\usepackage{subfig}
\usepackage{amssymb}
\usepackage{makecell}

\title{Cross-lingual Transfer or Machine Translation? \\
On Data Augmentation for Monolingual Semantic Textual Similarity}

\name{Sho Hoshino, Akihiko Kato, Soichiro Murakami, Peinan Zhang} 

\address{\includegraphics[width=0.15\linewidth]{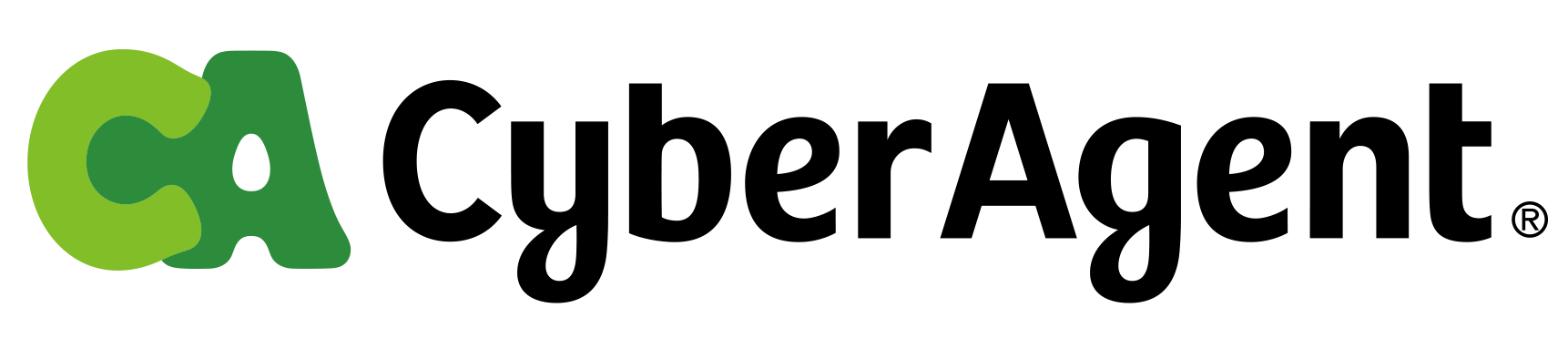} \\
         \{hoshino\_sho, kato\_akihiko, murakami\_soichiro, zhang\_peinan\}@cyberagent.co.jp\\}

\abstract{
Learning better sentence embeddings leads to improved performance for natural language understanding tasks including semantic textual similarity (STS) and natural language inference (NLI).
As prior studies leverage large-scale labeled NLI datasets for fine-tuning masked language models to yield sentence embeddings, task performance for languages other than English is often left behind.
In this study, we directly compared two data augmentation techniques as potential solutions for monolingual STS:
\begin{inparaenum}[(a)]
\item \textit{cross-lingual transfer} that exploits English resources alone as training data to yield non-English sentence embeddings as zero-shot inference, and
\item \textit{machine translation} that coverts English data into pseudo non-English training data in advance.
\end{inparaenum}
In our experiments on monolingual STS in Japanese and Korean, we find that the two data techniques yield performance on par.
Rather, we find a superiority of the Wikipedia domain over the NLI domain for these languages, in contrast to prior studies that focused on NLI as training data.
Combining our findings, we demonstrate that the cross-lingual transfer of Wikipedia data exhibits improved performance, and that native Wikipedia data can further improve performance for monolingual STS.
 \\ \newline  }

\begin{document}

\maketitleabstract

\section{Introduction}
Monolingual semantic textual similarity \citep[STS;][]{agirre-etal-2016-semeval} has been used as a progress milestone for the learning of sentence embeddings \cite{reimers-gurevych-2019-sentence,gao-etal-2021-simcse}.
Given two sentences in a target language, our task was to predict the similarity between the two sentences.
Monolingual STS is a core part of natural language understanding tasks \cite{wang-etal-2018-glue} and is related to natural language inference \citep[NLI;][]{10.1007/11736790_9} where the task was to predict whether one sentence entails another.
NLI data have been used for unsupervised STS without using any STS-specific data and are therefore considered to be suitable training data for monolingual STS.

Because such a quantity of training data is not always available in languages other than English, data augmentation techniques including
\begin{inparaenum}[(a)]
\item \textit{cross-lingual transfer} \cite{conneau-etal-2018-xnli,gogoulou-etal-2022-cross,wang-etal-2022-english} and
\item \textit{machine translation} \cite{conneau-etal-2018-xnli,ham-etal-2020-kornli,yanaka-mineshima-2022-compositional}
\end{inparaenum}
have been extensively applied.
However, aside from the monumental prior study on NLI \cite{conneau-etal-2018-xnli}, there have been few to no comprehensive studies that directly compared the two different data augmentation techniques particularly on monolingual STS, which differs from multilingual STS \cite{cer-etal-2017-semeval} in that the given two sentences are in the same language.
Therefore, we investigated the following research questions to search for the most suitable data augmentation technique for monolingual STS, which would have the greatest need for monolingual data:
\begin{inparadesc}
\item[RQ1] \textit{Between cross-lingual transfer or machine translation, which is better?}
\item[RQ2] \textit{Do these approaches yield performance on par with that of a state-of-the-art multilingual model?}
\end{inparadesc}

\begin{figure}[t]
\centering
\subfloat[cross-lingual transfer]{
\includegraphics[width=0.45\linewidth]{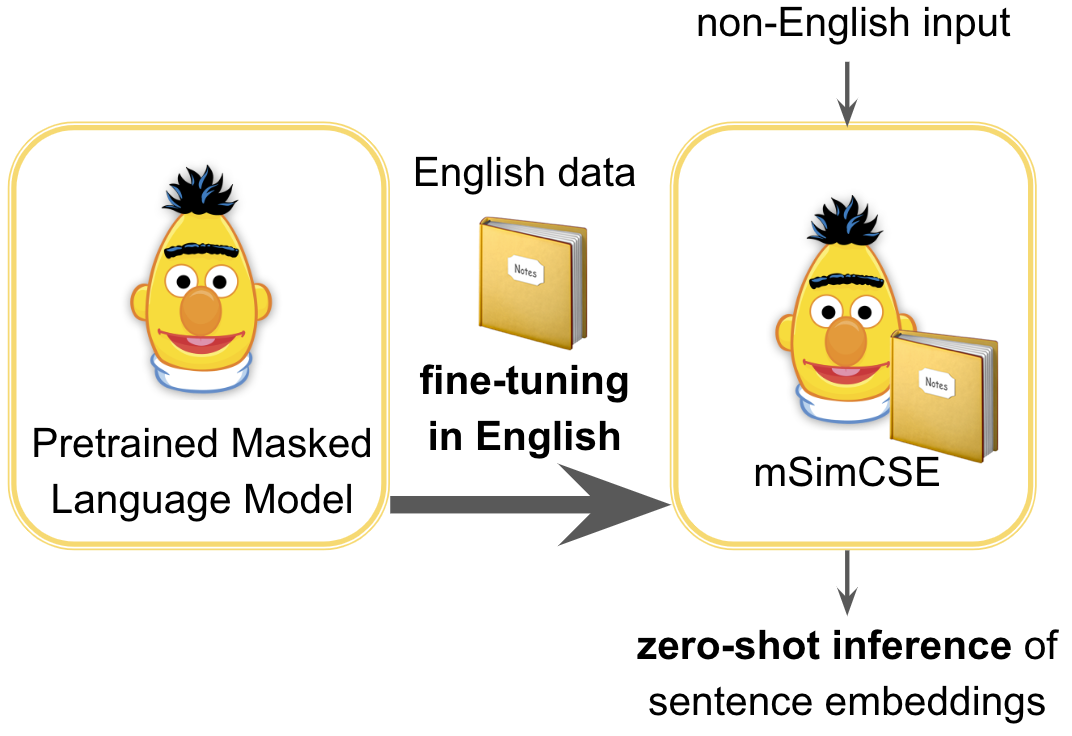}
}
\quad
\subfloat[machine translation]{
\includegraphics[width=0.45\linewidth]{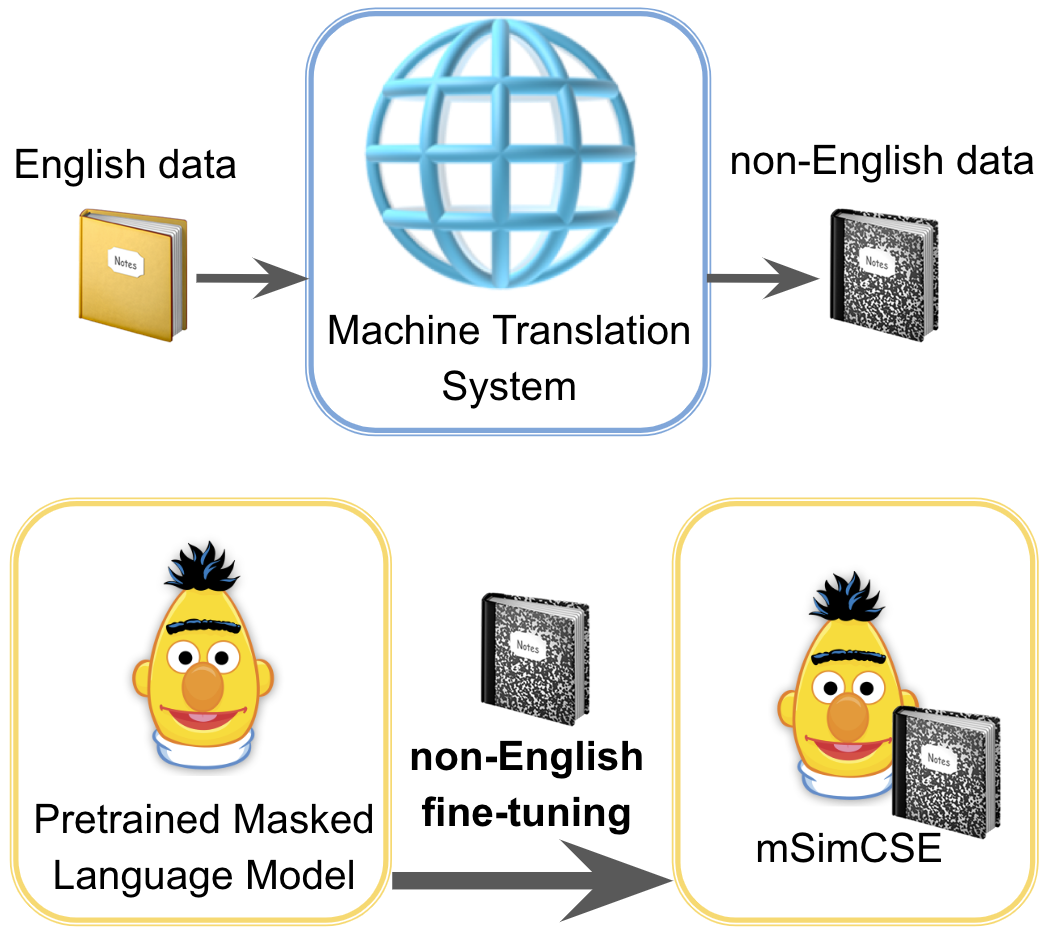}
}
\caption{Illustration of the two different data augmentation techniques applied from English to non-English.}\label{fig:eyecatch}
\end{figure}

In this study, we empirically evaluated the two different data augmentation techniques using Japanese and Korean, which can be seen as relatively low-resourced and linguistically dissimilar languages when compared with English, and thus are challenging for the two techniques.
We used unsupervised multilingual SimCSE \citep[mSimCSE;][]{wang-etal-2022-english} as our test bed, which is a multilingual extension of unsupervised SimCSE \cite{gao-etal-2021-simcse} that uses sentence-pair contrastive learning for self-supervised learning of unlabeled data.

Specifically, we trained mSimCSE models in two ways, each corresponding to the two data augmentation techniques:
\begin{inparaenum}[(a)]
\item by using English training data alone as cross-lingual transfer, and
\item by using machine-translated data from English into Korean and Japanese.
\end{inparaenum}
We also compared trained mSimCSE models with LaBSE \cite{feng-etal-2022-language}, a state-of-the-art multilingual model.

In our experiments on monolingual STS, we demonstrate that cross-lingual transfer achieves performance on par with that of machine translation (\textbf{RQ1}).
We further demonstrate that the combination of cross-lingual transfer and Wikipedia domain data exhibits the best performance outperforming or comparable to that of LaBSE (\textbf{RQ2}).
In contrast to prior studies, we also observed that Wikipedia domain data can be used as an alternative drop-in replacement for NLI domain data when used as unlabeled training data for monolingual STS.

Since Wikipedia domain data is easier to obtain than NLI domain data, our results suggest a recipe for training better sentence embeddings using a large-scale multilingual Wikipedia dataset.
As such a pilot study, we actually report improved results using a Japanese portion of the Wikipedia data.

\section{Data Augmentation Techniques}
Data augmentation \cite{feng-etal-2021-survey} is a generic strategy to deal with relatively low-resourced situations in a target language to increase the number of training examples (Figure~\ref{fig:eyecatch}).
We explain two different data augmentation techniques as follows.
\begin{asparadesc}
\item[Cross-lingual transfer] leverages training data in other languages but not in the target language.
We conduct fine-tuning of pretrained masked language models using the available data, without leveraging target language resources.\footnote
{In this context, the pretrained models must be capable of processing a target language and its vocabulary, unlike cross-lingual transfer to unseen languages \cite{artetxe-etal-2020-cross}.}
After that, we use the trained model to perform zero-shot inference in the target language.
This cost-effective approach has virtually no cost for obtaining data.
\citet{conneau-etal-2018-xnli} studied the performance of such cross-lingual transfer from English to other languages using a multilingual NLI dataset (XNLI).
\item[Machine translation] leverages the same data, but this time by using a machine translation system to translate it from other languages into the target language.
We then perform fine-tuning of pretrained masked language models using the machine-translated data.
After that, we use the trained model to perform normal inference in the target language, unlike cross-lingual transfer.
This pay-as-you-go approach has two variations.
\citet{conneau-etal-2018-xnli} created the training portion of the XNLI dataset using a neural machine translation system (\textsc{Translate Train}).
They also attained comparable results by translating test data only at runtime (\textsc{Translate Test}).
\end{asparadesc}

\section{Experiments}
We compared the performance of cross-lingual transfer and machine translation on unsupervised learning for monolingual STS \cite{agirre-etal-2016-semeval}.
Given two sentences in a target language, our task was to predict the similarity between the two sentences using unlabeled training data.
This task differs from supervised learning that exploits labeled data, because our focus is on the relatively low-resourced situations in Japanese and Korean.
The task also differs from multilingual STS in that the given two sentences are in the same language.

\subsection{Setup}

\begin{table}[t]
\centering
\footnotesize
\begin{tabular}{cc}
\toprule
Hyperparameter & Values \\
\midrule
\# Epochs & 1 \\
Learning rate & 5e-5 \\
Warmup rate & 10\% \\
Batch size & \{32, 64, 128\} \\
\bottomrule
\end{tabular}
\caption{
Hyperparameters.
}\label{tab:hyper}
\end{table}

\label{setup}
More specifically, we investigated the performance of sentence embeddings obtained as a result of unsupervised learning for monolingual STS.
Following mSimCSE \cite{wang-etal-2022-english}, we fine-tuned multilingual pretrained masked language models by feeding various training data as different data augmentation techniques.
Since XLM-R \cite{conneau-etal-2020-unsupervised} is used as a common baseline in the relevant literature, we used XLM-R as the multilingual base model and performed fine-tuning using unsupervised SimCSE \cite{gao-etal-2021-simcse}.
We compared our models with LaBSE \cite{feng-etal-2022-language} without fine-tuning as a strong fully supervised baseline.

We used random hardware available at Google Colaboratory\footnote{\url{https://colab.research.google.com/}} and Sentence-Transformers 2.2.0 \cite{reimers-gurevych-2019-sentence} for our implementation.\footnote
{The details of our implementation are described in the Appendix \ref{sec:implementation}, followed by the replication study of unsupervised SimCSE reported in the Appendix \ref{sec:replication}.}
We followed the hyperparameters used in Sentence-Transformers as best practices, unless listed in Table~\ref{tab:hyper}.

\subsection{Datasets}

\begin{table}[t]
\centering
\footnotesize
\begin{tabular}{@{}ccccc@{}}
\toprule
Name & Size & Domain & Lang. & Sources \\
\midrule
SNLI & 0.5M & NLI & en \\
MNLI & 0.5M & NLI & en \\
JSNLI & 0.5M & NLI & MT\textsubscript{ja} & SNLI \\
KorNLI & 1.0M & NLI & MT\textsubscript{ko} & SNLI, MNLI \\
Wiki & 1.0M & Wikipedia & en \\
Wiki-40B & 1.0M & Wikipedia & ja \\
\bottomrule
\end{tabular}
\caption{
The statistics of the various training datasets we used.
The size column reports the approximate number of training examples.
MT denotes pseudo data machine-translated from the corresponding resources shown in the last column.
}\label{tab:datasets}
\end{table}

\paragraph{Training}
Table~\ref{tab:datasets} summarizes various datasets in English, Japanese, and Korean that are primarily used as our training data.
Following prior studies on monolingual STS \citep{reimers-gurevych-2019-sentence,gao-etal-2021-simcse}, we used NLI datasets that contain premises, hypotheses, and their relationship labels such as entailment, contradiction, and neutral \cite{10.1007/11736790_9}.
Specifically, we used premises and hypotheses as training examples of unsupervised learning, while simply discarding NLI labels.
We also used English and Japanese Wikipedia.\footnote
{The details of data preprocessing are described in the Appendix \ref{sec:preprocess}.}

For fine-tuning in English, we used the Stanford Natural Language Inference corpus \citep[SNLI;][]{bowman-etal-2015-large} and the Multi-Genre Natural Language Inference corpus \citep[MNLI;][]{williams-etal-2018-broad} as well as English Wikipedia \citep[Wiki;][]{gao-etal-2021-simcse}.
To match the data size of NLI datasets with its Wikipedia counterpart (1 million sentences), we used them in two ways\footnote
{An ablation study with different data sizes is reported in the Appendix \ref{sec:ablation}.}:
\begin{inparaenum}[(i)]
\item only the premises portions of SNLI and MNLI (\textbf{NLI}; 942,854 sentences), and
\item both the premise and hypothesis portions of SNLI (\textbf{SNLI}; 1,100,304 sentences).
\end{inparaenum}

For fine-tuning in Japanese and Korean, machine-translated English datasets are used instead.
For instance, the JSNLI dataset \cite{jsnli} contains machine-translated SNLI and the KorNLI dataset \cite{ham-etal-2020-kornli} contains machine-translated SNLI and MNLI, respectively.
We also used Japanese Wikipedia data created from multilingual Wikipedia \citep[Wiki-40B;][]{guo-etal-2020-wiki}.

\paragraph{Evaluation}
We used the following labeled STS datasets as our evaluation data: STS-B \cite{cer-etal-2017-semeval} for English; JSICK-STS \cite{yanaka-mineshima-2022-compositional}, a human-translated SICK \cite{marelli-etal-2014-sick}, and JGLUE-JSTS \cite{kurihara-etal-2022-jglue} for Japanese; and KorSTS \cite{ham-etal-2020-kornli}, a human-translated STS-B \cite{cer-etal-2017-semeval}, and KLUE-STS \cite{park-etal-2021-klue} for Korean.

\subsection{Evaluation Protocol}
Following \citet{gao-etal-2021-simcse}, we evaluate sentence embeddings by measuring the correlation between the human labels $[0, 5]$ and the cosine distance between two given sentences, using Spearman's rank correlation coefficient \cite{spearman-1904}.
However, instead of using test portions of the labeled STS datasets, we used development portions as our evaluation data to avoid the unnecessary overfitting to English development data \cite{keung-etal-2020-dont}, which would caused by applying the standard evaluation protocol to cross-lingual transfer experiments.

\begin{table}[t]
\centering
\footnotesize
\begin{tabular}{@{}l@{}cccc@{}}
\toprule
& \multicolumn{2}{c}{Japanese} & \multicolumn{2}{c}{Korean} \\
\multicolumn{1}{c}{Models} & \makecell{JSICK-\\STS} & \makecell{JGLUE-\\JSTS} & KorSTS & \makecell{KLUE-\\STS} \\
\midrule
\multicolumn{5}{c}{\textit{Cross-lingual transfer}} \\
mSimCSE\textsubscript{en} \\
+ NLI & 79.55 & 74.12 & 74.68 & 72.87 \\
+ SNLI & \textbf{80.31} & 74.73 & 75.21 & 65.83 \\
\midrule
\multicolumn{5}{c}{\textit{Machine translation}} \\
mSimCSE\textsubscript{MT\textsubscript{ja}} \\
+ JSNLI & 78.41 & \textbf{75.55} \\
mSimCSE\textsubscript{MT\textsubscript{kr}} \\
+ KorNLI & & & \textbf{75.43} & \textbf{74.45} \\
\bottomrule
\end{tabular}
\caption{
Comparisons of cross-lingual transfer and machine translation as data augmentation.
We report Spearman’s correlation [\%].
Higher is better.
}\label{tab:controlled}
\end{table}

\begin{table}[t]
\centering
\footnotesize
\subfloat[Monolingual models.]{
\begin{tabular}{l@{}ccc}
\toprule
& English \\
\multicolumn{1}{c}{Models} & STS-B \\
\midrule
\multicolumn{2}{c}{\textit{Unsupervised}} \\
RoBERTa\textsubscript{large} & 56.29 \\
SimCSE \\
+ Wiki & \textbf{85.95} \\
\bottomrule
\end{tabular}
}
\quad
\subfloat[Multilingual models.]{
\begin{tabular}{l@{}ccc}
\toprule
& English \\
\multicolumn{1}{c}{Models} & STS-B \\
\midrule
\multicolumn{2}{c}{\textit{Unsupervised}} \\
XLM-R\textsubscript{large} & 43.54 \\
mSimCSE\textsubscript{en} \\
+ Wiki & \textbf{83.54} \\
+ NLI & 76.98 \\
\bottomrule
\end{tabular}
}
\caption{Comparisons of different data domains.
}\label{tab:domain}
\end{table}

\begin{table}[t]
\centering
\footnotesize
\begin{tabular}{@{}l@{}cccc@{}}
\toprule
& \multicolumn{2}{c}{Japanese} & \multicolumn{2}{c}{Korean} \\
\multicolumn{1}{c}{Models} & \makecell{JSICK-\\STS} & \makecell{JGLUE-\\JSTS} & KorSTS & \makecell{KLUE-\\STS} \\
\midrule
\multicolumn{5}{c}{\textit{Unsupervised}} \\
XLM-R\textsubscript{large} & 61.87 & 54.19 & 49.26 & 20.13 \\
mSimCSE\textsubscript{en} \\
+ Wiki & \textbf{81.02} & \textbf{77.62} & \textbf{80.38} & 81.42 \\
+ SNLI & 80.31 & 74.73 & 75.21 & 65.83 \\
\midrule
\midrule
\multicolumn{5}{c}{\textit{Fully supervised}} \\
LaBSE & 76.77 & 76.12 & 73.01 & \textbf{82.81} \\
\bottomrule
\end{tabular}
\caption{
Comparisons of the best-performing cross-lingual transfer of Wikipedia data and LaBSE.
}\label{tab:best}
\end{table}

\subsection{Results}
Table~\ref{tab:controlled} summarizes the comparisons of the two data augmentation techniques under a controlled setting within the same NLI domain.
In this fair setting, machine translation (mSimCSE\textsubscript{MT}) slightly outperformed cross-lingual transfer (mSimCSE\textsubscript{en}).
Specifically, in the case of Korean, mSimCSE\textsubscript{MT\textsubscript{kr}} almost outperformed mSimCSE\textsubscript{en}.
Similarly, in the case of Japanese, mSimCSE\textsubscript{MT\textsubscript{ja}} outperformed mSimCSE\textsubscript{en} with the exception on JSICK-STS.

However, we are yet to conclude that machine translation is better than cross-lingual transfer.
Table~\ref{tab:domain} summarizes the comparisons between NLI and Wikipedia domains in English.
Unlike prior studies focusing on NLI as training data, surprisingly, we obtained different results in this domain-aware setting.
Specifically, the mSimCSE\textsubscript{en} model trained using Wikipedia data outperformed its counterpart using NLI data, suggesting a superiority of Wikipedia as unlabeled training data.

These new results led us to the combination of our findings, namely the cross-lingual transfer of Wikipedia domain data.
Table~\ref{tab:best} summarizes the comparisons involving both the Wikipedia and NLI domains.
In this best-performing setting following \citet{gao-etal-2021-simcse}, the cross-lingual transfer of Wikpedia outperformed the machine translation of NLI, resulting in performance almost outperforming that of LaBSE.
Specifically, the mSimCSE\textsubscript{en} model trained using Wikipedia data outperformed LaBSE with the only exception on KLUE-STS.\footnote
{We also performed an analysis of the obtained sentence embeddings using KLUE-STS, which is reported in the Appendix \ref{sec:analysis}}

\section{Discussion}

\paragraph{Between cross-lingual transfer or machine translation, which is better?}
It depends.
On one hand, machine translation can outperform cross-lingual transfer if we used the same NLI domain as training data.
On the other hand, the cross-lingual transfer of Wikipedia domain data outperformed the machine-translated NLI domain data.
These results rather suggest the effectiveness of Wikipedia as unlabeled training data.

\paragraph{Do these approaches yield performance on par with that of a state-of-the-art multilingual model?}
Yes, we found that the cross-lingual transfer of Wikipedia data can outperform LaBSE.
This posed us a few more questions:
Should we pursue the direction of English as training data proxy, similar to mSimCSE?
Are there some benefit from using native multilingual data, if we could create it without using machine translation, similar to LaBSE?

\begin{table}[t]
\centering
\footnotesize
\begin{tabular}{l@{}ccc}
\toprule
& English & \multicolumn{2}{c}{Japanese} \\
\multicolumn{1}{c}{Models} & STS-B & \makecell{JSICK-\\STS} & \makecell{JGLUE-\\JSTS} \\
\midrule
\multicolumn{4}{c}{\textit{Cross-lingual transfer}} \\
mSimCSE\textsubscript{en} \\
+ Wiki & \textbf{83.54} & 81.02 & 77.62 \\
\midrule
\multicolumn{4}{c}{\textit{Native data}} \\
mSimCSE\textsubscript{ja} \\
+ Wiki-40B & 81.89 & \textbf{81.05} & \textbf{78.71} \\
\bottomrule
\end{tabular}
\caption{
Comparisons of English Wikipedia data cross-lingual transfer and native Japanese data.
}\label{tab:native}
\end{table}

Table~\ref{tab:native} summarizes additional results using native Japanese Wikipedia data.
This pilot study suggests that using native data can indeed improve performance over cross-lingual transfer of English data.
However, we are yet to answer the ultimate question of English as training data proxy.

\section{Related Work}

\paragraph{Learning Sentence Embeddings}
Sentence embeddings are learned representations of sentences within a single dense matrix, unlike word embeddings, which are represented in multiple dense matrices.
Several methods for fine-tuning pretrained masked language models have been proposed, including
\begin{inparaenum}[(a)]
\item SBERT \cite{reimers-gurevych-2019-sentence} using sentence-pair regression for supervised learning of labeled STS data, and
\item unsupervised SimCSE \cite{gao-etal-2021-simcse} using sentence-pair contrastive learning for self-supervised learning of unlabeled data.
\end{inparaenum}
Both methods yield good sentence embeddings in terms of cosine distance in STS.

\paragraph{Cross-lingual Transferability of Multilingual Models}
\citet{artetxe-schwenk-2019-massively} proposed LASER, language-agnostic representations of sentence embeddings learned from dedicated parallel data.
They studied cross-lingual transferability of LASER on XNLI \cite{conneau-etal-2018-xnli} and found improved performance in various languages.
\citet{feng-etal-2022-language} proposed LaBSE, which outperformed LASER in downstream tasks by using additional monolingual data with parallel data.
\citet{wang-etal-2022-english} proposed mSimCSE, a multilingual extension of SimCSE, which is perhaps the most related work.
They investigated cross-lingual transferability of mSimCSE by using various tasks including multilingual STS.

\paragraph{Cross-lingual Transferability of Monolingual Models}
\citet{artetxe-etal-2020-cross} studied cross-lingual transferability from one language to unseen languages.
They showed that the transfer learning of monolingual BERT \cite{devlin-etal-2019-bert} at the lexical level outperformed multilingual BERT on XQuAD.
\citet{reimers-gurevych-2020-making} studied transfer learning of SBERT at the sentence level, using multilingual knowledge distillation in English and Korean as monolingual and multilingual STS.
They observed performance better than that of monolingual fine-tuning in Korean.
\citet{gogoulou-etal-2022-cross} studied cross-lingual transferability from non-English languages into English by using various GLUE tasks \cite{wang-etal-2018-glue} including monolingual STS.

\section{Conclusion}
In this study, we empirically compared cross-lingual transfer and machine translation in terms of monolingual STS performance.
We chose Japanese and Korean as our test bed, because these languages are relatively low-resourced and linguistically dissimilar compared with English, and thus are challenging for the two data augmentation techniques.
We found that the cross-lingual transfer exhibits performance comparable to that of the machine translation depending on data domain.
We also found that, in contrast to prior studies, cross-lingual transfer of Wikipedia data achieved the best performance, outperforming or comparable to that of the state-of-the-art LaBSE.
Our future work will include fine-grained analysis of which types of data are better suitable for monolingual STS.

\section*{Limitations}
Our study focuses primarily on Japanese and Korean, which are often considered high-resource or mid-resource languages, but, at the same time, are relatively low-resourced with respect to STS training data and therefore suitable for data augmentation.
For this reason, our results are not directly applicable to other low-resource and regional languages, as even human-labeled STS evaluation data are lacking.
In addition, we did not conduct experiments using Wikipedia data machine-translated from English into Japanese and Korean, which will be part of our future work.

\section{Bibliographical References}\label{sec:reference}

\bibliography{custom}

\begin{thebibliography}{32}
\expandafter\ifx\csname natexlab\endcsname\relax\def\natexlab#1{#1}\fi

\bibitem[{Agirre et~al.(2016)Agirre, Banea, Cer, Diab, Gonzalez-Agirre, Mihalcea, Rigau, and Wiebe}]{agirre-etal-2016-semeval}
Eneko Agirre, Carmen Banea, Daniel Cer, Mona Diab, Aitor Gonzalez-Agirre, Rada Mihalcea, German Rigau, and Janyce Wiebe. 2016.
\newblock \href {https://doi.org/10.18653/v1/S16-1081} {{S}em{E}val-2016 task 1: Semantic textual similarity, monolingual and cross-lingual evaluation}.
\newblock In \emph{Proceedings of the 10th International Workshop on Semantic Evaluation}, pages 497--511.

\bibitem[{Artetxe et~al.(2020)Artetxe, Ruder, and Yogatama}]{artetxe-etal-2020-cross}
Mikel Artetxe, Sebastian Ruder, and Dani Yogatama. 2020.
\newblock \href {https://doi.org/10.18653/v1/2020.acl-main.421} {On the cross-lingual transferability of monolingual representations}.
\newblock In \emph{Proceedings of the 58th Annual Meeting of the Association for Computational Linguistics}, pages 4623--4637.

\bibitem[{Artetxe and Schwenk(2019)}]{artetxe-schwenk-2019-massively}
Mikel Artetxe and Holger Schwenk. 2019.
\newblock \href {https://doi.org/10.1162/tacl_a_00288} {Massively multilingual sentence embeddings for zero-shot cross-lingual transfer and beyond}.
\newblock \emph{Transactions of the Association for Computational Linguistics}, 7:597--610.

\bibitem[{Bowman et~al.(2015)Bowman, Angeli, Potts, and Manning}]{bowman-etal-2015-large}
Samuel~R. Bowman, Gabor Angeli, Christopher Potts, and Christopher~D. Manning. 2015.
\newblock \href {https://doi.org/10.18653/v1/D15-1075} {A large annotated corpus for learning natural language inference}.
\newblock In \emph{Proceedings of the 2015 Conference on Empirical Methods in Natural Language Processing}, pages 632--642.

\bibitem[{Cer et~al.(2017)Cer, Diab, Agirre, Lopez-Gazpio, and Specia}]{cer-etal-2017-semeval}
Daniel Cer, Mona Diab, Eneko Agirre, I{\~n}igo Lopez-Gazpio, and Lucia Specia. 2017.
\newblock \href {https://doi.org/10.18653/v1/S17-2001} {{SemEval}-2017 task 1: Semantic textual similarity multilingual and crosslingual focused evaluation}.
\newblock In \emph{Proceedings of the 11th International Workshop on Semantic Evaluation}, pages 1--14.

\bibitem[{Conneau et~al.(2020)Conneau, Khandelwal, Goyal, Chaudhary, Wenzek, Guzm{\'a}n, Grave, Ott, Zettlemoyer, and Stoyanov}]{conneau-etal-2020-unsupervised}
Alexis Conneau, Kartikay Khandelwal, Naman Goyal, Vishrav Chaudhary, Guillaume Wenzek, Francisco Guzm{\'a}n, Edouard Grave, Myle Ott, Luke Zettlemoyer, and Veselin Stoyanov. 2020.
\newblock \href {https://doi.org/10.18653/v1/2020.acl-main.747} {Unsupervised cross-lingual representation learning at scale}.
\newblock In \emph{Proceedings of the 58th Annual Meeting of the Association for Computational Linguistics}, pages 8440--8451.

\bibitem[{Conneau et~al.(2018)Conneau, Rinott, Lample, Williams, Bowman, Schwenk, and Stoyanov}]{conneau-etal-2018-xnli}
Alexis Conneau, Ruty Rinott, Guillaume Lample, Adina Williams, Samuel Bowman, Holger Schwenk, and Veselin Stoyanov. 2018.
\newblock \href {https://doi.org/10.18653/v1/D18-1269} {{XNLI}: Evaluating cross-lingual sentence representations}.
\newblock In \emph{Proceedings of the 2018 Conference on Empirical Methods in Natural Language Processing}, pages 2475--2485.

\bibitem[{Dagan et~al.(2005)Dagan, Glickman, and Magnini}]{10.1007/11736790_9}
Ido Dagan, Oren Glickman, and Bernardo Magnini. 2005.
\newblock \href {https://doi.org/10.1007/11736790_9} {The {PASCAL} recognising textual entailment challenge}.
\newblock In \emph{Proceedings of the 1st {PASCAL} Machine Learning Challenges Workshop}, pages 177--190.

\bibitem[{Devlin et~al.(2019)Devlin, Chang, Lee, and Toutanova}]{devlin-etal-2019-bert}
Jacob Devlin, Ming-Wei Chang, Kenton Lee, and Kristina Toutanova. 2019.
\newblock \href {https://doi.org/10.18653/v1/N19-1423} {{BERT}: Pre-training of deep bidirectional transformers for language understanding}.
\newblock In \emph{Proceedings of the 2019 Conference of the North {A}merican Chapter of the Association for Computational Linguistics: Human Language Technologies}, pages 4171--4186.

\bibitem[{Feng et~al.(2022)Feng, Yang, Cer, Arivazhagan, and Wang}]{feng-etal-2022-language}
Fangxiaoyu Feng, Yinfei Yang, Daniel Cer, Naveen Arivazhagan, and Wei Wang. 2022.
\newblock \href {https://aclanthology.org/2022.acl-long.62} {Language-agnostic {BERT} sentence embedding}.
\newblock In \emph{Proceedings of the 60th Annual Meeting of the Association for Computational Linguistics}, pages 878--891.

\bibitem[{Feng et~al.(2021)Feng, Gangal, Wei, Chandar, Vosoughi, Mitamura, and Hovy}]{feng-etal-2021-survey}
Steven~Y. Feng, Varun Gangal, Jason Wei, Sarath Chandar, Soroush Vosoughi, Teruko Mitamura, and Eduard Hovy. 2021.
\newblock \href {https://doi.org/10.18653/v1/2021.findings-acl.84} {A survey of data augmentation approaches for {NLP}}.
\newblock In \emph{Findings of the Association for Computational Linguistics: {ACL-IJCNLP} 2021}, pages 968--988.

\bibitem[{Gao et~al.(2021)Gao, Yao, and Chen}]{gao-etal-2021-simcse}
Tianyu Gao, Xingcheng Yao, and Danqi Chen. 2021.
\newblock \href {https://doi.org/10.18653/v1/2021.emnlp-main.552} {{SimCSE}: Simple contrastive learning of sentence embeddings}.
\newblock In \emph{Proceedings of the 2021 Conference on Empirical Methods in Natural Language Processing}, pages 6894--6910.

\bibitem[{Gogoulou et~al.(2022)Gogoulou, Ekgren, Isbister, and Sahlgren}]{gogoulou-etal-2022-cross}
Evangelia Gogoulou, Ariel Ekgren, Tim Isbister, and Magnus Sahlgren. 2022.
\newblock \href {https://aclanthology.org/2022.lrec-1.100} {Cross-lingual transfer of monolingual models}.
\newblock In \emph{Proceedings of the 13th International Conference on Language Resources and Evaluation}, pages 948--955.

\bibitem[{Guo et~al.(2020)Guo, Dai, Vrande{\v{c}}i{\'c}, and Al-Rfou}]{guo-etal-2020-wiki}
Mandy Guo, Zihang Dai, Denny Vrande{\v{c}}i{\'c}, and Rami Al-Rfou. 2020.
\newblock \href {https://aclanthology.org/2020.lrec-1.297} {{W}iki-40{B}: Multilingual language model dataset}.
\newblock In \emph{Proceedings of the Twelfth Language Resources and Evaluation Conference}, pages 2440--2452.

\bibitem[{Ham et~al.(2020)Ham, Choe, Park, Choi, and Soh}]{ham-etal-2020-kornli}
Jiyeon Ham, Yo~Joong Choe, Kyubyong Park, Ilji Choi, and Hyungjoon Soh. 2020.
\newblock \href {https://doi.org/10.18653/v1/2020.findings-emnlp.39} {{KorNLI} and {KorSTS}: New benchmark datasets for {K}orean natural language understanding}.
\newblock In \emph{Findings of the Association for Computational Linguistics: {EMNLP} 2020}, pages 422--430.

\bibitem[{Keung et~al.(2020)Keung, Lu, Salazar, and Bhardwaj}]{keung-etal-2020-dont}
Phillip Keung, Yichao Lu, Julian Salazar, and Vikas Bhardwaj. 2020.
\newblock \href {https://doi.org/10.18653/v1/2020.emnlp-main.40} {Don{'}t use {E}nglish dev: On the zero-shot cross-lingual evaluation of contextual embeddings}.
\newblock In \emph{Proceedings of the 2020 Conference on Empirical Methods in Natural Language Processing}, pages 549--554.

\bibitem[{Kudo and Richardson(2018)}]{kudo-richardson-2018-sentencepiece}
Taku Kudo and John Richardson. 2018.
\newblock \href {https://doi.org/10.18653/v1/D18-2012} {{SentencePiece}: A simple and language independent subword tokenizer and detokenizer for neural text processing}.
\newblock In \emph{Proceedings of the 2018 Conference on Empirical Methods in Natural Language Processing: System Demonstrations}, pages 66--71.

\bibitem[{Kurihara et~al.(2022)Kurihara, Kawahara, and Shibata}]{kurihara-etal-2022-jglue}
Kentaro Kurihara, Daisuke Kawahara, and Tomohide Shibata. 2022.
\newblock \href {https://aclanthology.org/2022.lrec-1.317} {{JGLUE}: {J}apanese general language understanding evaluation}.
\newblock In \emph{Proceedings of the 13th Language Resources and Evaluation Conference}, pages 2957--2966.

\bibitem[{Liu et~al.(2019)Liu, Ott, Goyal, Du, Joshi, Chen, Levy, Lewis, Zettlemoyer, and Stoyanov}]{liu-etal-2019-roberta}
Yinhan Liu, Myle Ott, Naman Goyal, Jingfei Du, Mandar Joshi, Danqi Chen, Omer Levy, Mike Lewis, Luke Zettlemoyer, and Veselin Stoyanov. 2019.
\newblock \href {https://doi.org/10.48550/ARXIV.1907.11692} {{RoBERTa}: A robustly optimized {BERT} pretraining approach}.
\newblock Preprint, arXiv:1907.11692.

\bibitem[{Marelli et~al.(2014)Marelli, Menini, Baroni, Bentivogli, Bernardi, and Zamparelli}]{marelli-etal-2014-sick}
Marco Marelli, Stefano Menini, Marco Baroni, Luisa Bentivogli, Raffaella Bernardi, and Roberto Zamparelli. 2014.
\newblock \href {https://aclanthology.org/L14-1314/} {A {SICK} cure for the evaluation of compositional distributional semantic models}.
\newblock In \emph{Proceedings of the 9th International Conference on Language Resources and Evaluation}, pages 216--223.

\bibitem[{Park et~al.(2021)Park, Moon, Kim, Cho, Han, Park, Song, Kim, Song, Oh, Lee, Oh, Lyu, Jeong, Lee, Seo, Lee, Kim, Lee, Jang, Do, Kim, Lim, Lee, Park, Shin, Kim, Park, Oh, Ha, and Cho}]{park-etal-2021-klue}
Sungjoon Park, Jihyung Moon, Sungdong Kim, Won~Ik Cho, Ji~Yoon Han, Jangwon Park, Chisung Song, Junseong Kim, Youngsook Song, Taehwan Oh, Joohong Lee, Juhyun Oh, Sungwon Lyu, Younghoon Jeong, Inkwon Lee, Sangwoo Seo, Dongjun Lee, Hyunwoo Kim, Myeonghwa Lee, Seongbo Jang, Seungwon Do, Sunkyoung Kim, Kyungtae Lim, Jongwon Lee, Kyumin Park, Jamin Shin, Seonghyun Kim, Lucy Park, Alice Oh, Jung-Woo Ha, and Kyunghyun Cho. 2021.
\newblock \href {https://datasets-benchmarks-proceedings.neurips.cc/paper/2021/file/98dce83da57b0395e163467c9dae521b-Paper-round2.pdf} {{KLUE}: Korean language understanding evaluation}.
\newblock In \emph{Proceedings of the Neural Information Processing Systems Track on Datasets and Benchmarks}.

\bibitem[{Reimers and Gurevych(2019)}]{reimers-gurevych-2019-sentence}
Nils Reimers and Iryna Gurevych. 2019.
\newblock \href {https://doi.org/10.18653/v1/D19-1410} {Sentence-{BERT}: Sentence embeddings using {S}iamese {BERT}-networks}.
\newblock In \emph{Proceedings of the 2019 Conference on Empirical Methods in Natural Language Processing and the 9th International Joint Conference on Natural Language Processing}, pages 3982--3992.

\bibitem[{Reimers and Gurevych(2020)}]{reimers-gurevych-2020-making}
Nils Reimers and Iryna Gurevych. 2020.
\newblock \href {https://doi.org/10.18653/v1/2020.emnlp-main.365} {Making monolingual sentence embeddings multilingual using knowledge distillation}.
\newblock In \emph{Proceedings of the 2020 Conference on Empirical Methods in Natural Language Processing}, pages 4512--4525.

\bibitem[{Spearman(1904)}]{spearman-1904}
Charles Spearman. 1904.
\newblock \href {http://www.jstor.org/stable/1412159} {The proof and measurement of association between two things}.
\newblock \emph{The American Journal of Psychology}, 15(1):72--101.

\bibitem[{van~der Maaten and Hinton(2008)}]{maaten-hinton-2008-tsne}
Laurens van~der Maaten and Geoffrey Hinton. 2008.
\newblock \href {http://jmlr.org/papers/v9/vandermaaten08a.html} {Visualizing data using {t-SNE}}.
\newblock \emph{Journal of Machine Learning Research}, 9(86):2579--2605.

\bibitem[{Wang et~al.(2018)Wang, Singh, Michael, Hill, Levy, and Bowman}]{wang-etal-2018-glue}
Alex Wang, Amanpreet Singh, Julian Michael, Felix Hill, Omer Levy, and Samuel Bowman. 2018.
\newblock \href {https://doi.org/10.18653/v1/W18-5446} {{GLUE}: A multi-task benchmark and analysis platform for natural language understanding}.
\newblock In \emph{Proceedings of the 2018 {EMNLP} Workshop {B}lackbox{NLP}}, pages 353--355.

\bibitem[{Wang and Isola(2020)}]{pmlr-v119-wang20k}
Tongzhou Wang and Phillip Isola. 2020.
\newblock \href {https://proceedings.mlr.press/v119/wang20k.html} {Understanding contrastive representation learning through alignment and uniformity on the hypersphere}.
\newblock In \emph{Proceedings of the 37th International Conference on Machine Learning}, pages 9929--9939.

\bibitem[{Wang et~al.(2022)Wang, Wu, and Neubig}]{wang-etal-2022-english}
Yaushian Wang, Ashley Wu, and Graham Neubig. 2022.
\newblock \href {https://aclanthology.org/2022.emnlp-main.621} {{E}nglish contrastive learning can learn universal cross-lingual sentence embeddings}.
\newblock In \emph{Proceedings of the 2022 Conference on Empirical Methods in Natural Language Processing}, pages 9122--9133.

\bibitem[{Williams et~al.(2018)Williams, Nangia, and Bowman}]{williams-etal-2018-broad}
Adina Williams, Nikita Nangia, and Samuel Bowman. 2018.
\newblock \href {https://doi.org/10.18653/v1/N18-1101} {A broad-coverage challenge corpus for sentence understanding through inference}.
\newblock In \emph{Proceedings of the 2018 Conference of the North {A}merican Chapter of the Association for Computational Linguistics: Human Language Technologies}, pages 1112--1122.

\bibitem[{Wolf et~al.(2020)Wolf, Debut, Sanh, Chaumond, Delangue, Moi, Cistac, Rault, Louf, Funtowicz, Davison, Shleifer, von Platen, Ma, Jernite, Plu, Xu, Le~Scao, Gugger, Drame, Lhoest, and Rush}]{wolf-etal-2020-transformers}
Thomas Wolf, Lysandre Debut, Victor Sanh, Julien Chaumond, Clement Delangue, Anthony Moi, Pierric Cistac, Tim Rault, Remi Louf, Morgan Funtowicz, Joe Davison, Sam Shleifer, Patrick von Platen, Clara Ma, Yacine Jernite, Julien Plu, Canwen Xu, Teven Le~Scao, Sylvain Gugger, Mariama Drame, Quentin Lhoest, and Alexander Rush. 2020.
\newblock \href {https://doi.org/10.18653/v1/2020.emnlp-demos.6} {Transformers: State-of-the-art natural language processing}.
\newblock In \emph{Proceedings of the 2020 Conference on Empirical Methods in Natural Language Processing: System Demonstrations}, pages 38--45.

\bibitem[{Yanaka and Mineshima(2022)}]{yanaka-mineshima-2022-compositional}
Hitomi Yanaka and Koji Mineshima. 2022.
\newblock \href {https://doi.org/10.1162/tacl_a_00518} {Compositional evaluation on {J}apanese textual entailment and similarity}.
\newblock \emph{Transactions of the Association for Computational Linguistics}, 10:1266--1284.

\bibitem[{Yoshikoshi et~al.(2020)Yoshikoshi, Kawahara, and Kurohashi}]{jsnli}
Takumi Yoshikoshi, Daisuke Kawahara, and Sadao Kurohashi. 2020.
\newblock \href {https://nlp.ist.i.kyoto-u.ac.jp/?\%E6\%97\%A5\%E6\%9C\%AC\%E8\%AA\%9ESNLI\%28JSNLI\%29\%E3\%83\%87\%E3\%83\%BC\%E3\%82\%BF\%E3\%82\%BB\%E3\%83\%83\%E3\%83\%88} {Multilingualization of a natural language inference dataset using machine translation}.
\newblock \emph{{IPSJ SIG} Technical Report}, NL-244(6):1--8.
\newblock In Japanese.

\end{thebibliography}
\bibliographystyle{lrec-coling2024-natbib}

\clearpage
\appendix

\section{Implementation Details}
\label{sec:implementation}

In this study, we reimplemented the unsupervised SimCSE \cite{gao-etal-2021-simcse} instead of utilizing the original implementation provided by the authors\footnote{\url{https://github.com/princeton-nlp/SimCSE}}.
We used Sentence-Transformers 2.2.0 \cite{reimers-gurevych-2019-sentence}, Hugging Face Transformers 4.18.0 \cite{wolf-etal-2020-transformers}, and SentencePiece 0.1.96 \cite{kudo-richardson-2018-sentencepiece}.
We also used the publicly available models of XLM-R\footnote{\url{https://huggingface.co/xlm-roberta-large}}, LaBSE\footnote{\url{https://huggingface.co/sentence-transformers/LaBSE}}, RoBERTa\footnote{\url{https://huggingface.co/roberta-large}}, KLUE-RoBERTa\footnote{\url{https://huggingface.co/klue/roberta-large}}, and SBERT\footnote{\url{https://huggingface.co/sentence-transformers/roberta-large-nli-mean-tokens}}, as well as the unsupervised SimCSE models distributed by the original authors\footnote{\url{https://huggingface.co/princeton-nlp/unsup-simcse-roberta-large}}.
We report results obtained in a single run with the random seed fixed, while freezing hyperparameters for the sake of reproducibility \cite{keung-etal-2020-dont}.

\section{Data Preprocessing}
\label{sec:preprocess}

Here, we describe the details of our data preprocessing to make the data used in this study reproducible.
We have removed empty lines from our NLI and Wikipedia datasets and applied additional preprocessing as follows.

\paragraph{NLI datasets}
For JSNLI, we roughly detokenized the already-tokenized dataset by applying NFKC normalization and eliminating white spaces.
For JSICK-STS, we used its test portion to make it comparable with the other datasets.

\paragraph{Wikipedia datasets}
For English Wikipedia, we used the data\footnote{\url{https://huggingface.co/datasets/princeton-nlp/datasets-for-simcse}} distributed with the original implementation of SimCSE \cite{gao-etal-2021-simcse}, which contains exactly 1 million sentences but the authors did not report its details.

Therefore, we created our own Japanese Wikipedia data from the multilingual Wiki-40B dataset\footnote{\url{https://huggingface.co/datasets/wiki40b}} \cite{guo-etal-2020-wiki} as a replication study.
Specifically, we extracted Japanese paragraphs from Wiki-40B and applied sentence splitting using new line symbols and Japanese period markers.
To match the data size of the Japanese Wikipedia data with its English counterpart, we used only the first 1 million sentences.

\begin{figure}[t]
\centering
\begin{minipage}{.5\textwidth}
\subfloat[KLUE-RoBERTa\textsubscript{large}]{\label{fig:roberta}
\includegraphics[width=0.42\linewidth]{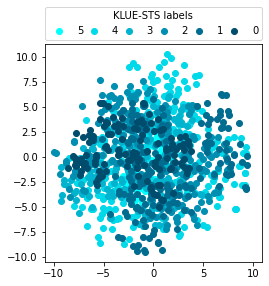}
}
\subfloat[SimCSE\textsubscript{MT\textsubscript{kr}}+KorNLI]{\label{fig:roberta-simcse}
\includegraphics[width=0.42\linewidth]{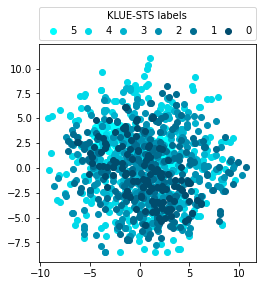}
}
\quad
\subfloat[XLM-R\textsubscript{large}]{\label{fig:xlmr}
\includegraphics[width=0.42\linewidth]{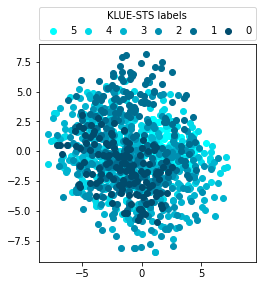}
}
\subfloat[mSimCSE\textsubscript{en} + Wiki]{\label{fig:xlmr-simcse}
\includegraphics[width=0.42\linewidth]{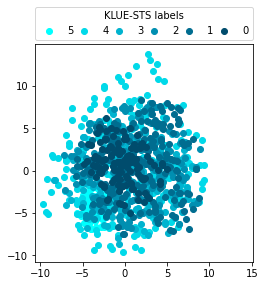}
}
\end{minipage}
\caption{
2D visualization of sentence embeddings on KLUE-STS.
Different scales are used to better illustrate the isotropic nature of SimCSE fine-tuning.
}\label{fig:visual}
\end{figure}

\section{Analysis of Sentence Embeddings}
\label{sec:analysis}

We performed an analysis of the obtained sentence embeddings using visualization and metrics.

\paragraph{Visualization}
Figure~\ref{fig:visual} provides a visualization of the sentence embeddings on KLUE-STS, which were obtained by applying t-distributed stochastic neighbor embedding \citep[t-SNE;][]{maaten-hinton-2008-tsne} for two-dimensional reduction.
We compared cross-lingual transfer (\ref{fig:xlmr-simcse}) with a Korean model (\ref{fig:roberta-simcse}), which was fine-tuned using KLUE-RoBERTa \cite{liu-etal-2019-roberta,park-etal-2021-klue}, unsupervised SimCSE \cite{gao-etal-2021-simcse}, and KorNLI.
We observe that cross-lingual transfer can maintain the isotropic nature of SimCSE fine-tuning.

\paragraph{Alignment and Uniformity}
We further confirmed the aforementioned trend by using alignment and uniformity metrics \cite{pmlr-v119-wang20k} to quantify the isotropic nature of SimCSE fine-tuning.
The monolingual model (\ref{fig:roberta-simcse}) scored $\ell_\mathrm{align}=0.3540$ and $\ell_\mathrm{uniform}=-3.3486$, while cross-lingual transfer (\ref{fig:xlmr-simcse}) scored similar values of $\ell_\mathrm{align}=0.2684$ and $\ell_\mathrm{uniform}=-3.2484$.

\begin{table}[t]
\centering
\footnotesize
\subfloat[Monolingual models.]{\label{tab:monolingual}
\begin{tabular}{cccc}
\toprule
& English \\
Models & STS-B \\
\midrule
\multicolumn{2}{c}{\textit{Unsupervised}} \\
RoBERTa\textsubscript{base} & 64.70 \\
SimCSE\textsubscript{orig} \\
\multicolumn{1}{l}{+ Wiki} & 84.26 \\
SimCSE\textsubscript{repro} \\
\multicolumn{1}{l}{+ Wiki} & 83.90 \\
\midrule
RoBERTa\textsubscript{large} & 56.29 \\
SimCSE\textsubscript{orig} \\
\multicolumn{1}{l}{+ Wiki} & 85.52 \\
SimCSE\textsubscript{repro} \\
\multicolumn{1}{l}{+ Wiki} & \textbf{85.95} \\
\midrule
\midrule
\multicolumn{2}{c}{\textit{English NLI Supervised}} \\
SBERT\textsubscript{base} \\
\multicolumn{1}{l}{+ NLI} & 80.73 \\
\midrule
SBERT\textsubscript{large} \\
\multicolumn{1}{l}{+ NLI} & 82.51 \\
\bottomrule
\end{tabular}
}
\subfloat[Multilingual models.]{\label{tab:multilingual}
\begin{tabular}{cccc}
\toprule
& English \\
Models & STS-B \\
\midrule
\multicolumn{2}{c}{\textit{Unsupervised}} \\
XLM-R\textsubscript{base} & 53.62 \\
mSimCSE\textsubscript{en} \\
\multicolumn{1}{l}{+ Wiki} & 76.44 \\
\multicolumn{1}{l}{+ NLI} & 75.79 \\
\midrule
XLM-R\textsubscript{large} & 43.54 \\
mSimCSE\textsubscript{en} \\
\multicolumn{1}{l}{+ Wiki} & \textbf{83.54} \\
\multicolumn{1}{l}{+ NLI} & 76.98 \\
\midrule
\midrule
\multicolumn{2}{c}{\textit{Fully supervised}} \\
LaBSE & 74.13 \\
\bottomrule
\end{tabular}
}
\caption{Comparison of unsupervised SimCSE in English on different training data and base models.
SimCSE\textsubscript{orig} denotes the models distributed by the original authors, whereas SimCSE\textsubscript{repro} and mSimCSE denote our implementation used in this study.
}\label{tab:replication}
\end{table}

\section{Replication Study}
\label{sec:replication}

We conducted a replication study of unsupervised SimCSE in English \cite{gao-etal-2021-simcse} by using various training data that have been extensively used in the relevant literature but not directly compared in the unsupervised SimCSE setup.
We used RoBERTa \cite{liu-etal-2019-roberta} as the monolingual base model and performed monolingual fine-tuning using unsupervised SimCSE.

As summarized in Table~\ref{tab:replication}, our implementation (SimCSE\textsubscript{repro} and mSimCSE) exhibited performance comparable to that of the models distributed by the original authors (SimCSE\textsubscript{orig}).

\section{Ablation Study}

\label{sec:ablation}
We conducted an ablation study on different sizes of training data in English, Japanese, and Korean.
In this comparison, we used NLI datasets, combining them in two ways:
\begin{inparaenum}[(a)]
\item only the premise portion, and
\item both the premise and hypothesis portions.
\end{inparaenum}
As a result, our data sizes varied from roughly 0.5M examples to up to 2.0M examples.

Table~\ref{tab:ablation} summarizes the results of the ablation study.
Our findings are twofold.
\begin{inparaenum}[(i)]
\item There was a minimum practical size of 1.0M examples.
All models trained using lower sizes suffered serious performance degradation.
\item We also observed that when the data size was already close to 1.0M examples, the case using the premise portion alone (\ref{tab:premise}) outperformed that using both the premise and hypothesis (\ref{tab:hypothesis}).
This result is convincing, as the hypothesis portion is artificially created from the premise portion \cite{10.1007/11736790_9} and there is not much variation in the relation between them.
\end{inparaenum}

\begin{table*}[t]
\centering
\footnotesize
\subfloat[The premise alone.]{\label{tab:premise}
\begin{tabular}{l@{}ccccc}
\toprule
& & English & Japanese & \multicolumn{2}{c}{Korean} \\
\multicolumn{1}{c}{Models} & Size & STS-B & \makecell{JGLUE-\\JSTS} & KorSTS & \makecell{KLUE-\\STS} \\
\midrule
\multicolumn{6}{c}{\textit{Unsupervised}} \\
XLM-R\textsubscript{base} & & 53.62 & 59.28 & 57.98 & 30.69 \\
mSimCSE\textsubscript{en} \\
\multicolumn{1}{l}{+ SNLI} & 0.5M & 72.61 & 71.56 & 69.52 & 52.13 \\
\multicolumn{1}{l}{+ NLI} & 1.0M & 75.79 & 71.83 & 73.61 & 60.66 \\
mSimCSE\textsubscript{MT\textsubscript{ja}} \\
\multicolumn{1}{l}{+ JSNLI} & 0.5M & 72.87 & 71.16 & 73.48 & 63.69 \\
mSimCSE\textsubscript{MT\textsubscript{kr}} \\
\multicolumn{1}{l}{+ KorNLI} & 1.0M & 75.39 & 73.02 & 73.34 & 67.13 \\
\midrule
XLM-R\textsubscript{large} & & 43.54 & 54.19 & 49.26 & 20.13 \\
mSimCSE\textsubscript{en} \\
\multicolumn{1}{l}{+ NLI} & 1.0M & 78.66 & 74.12 & 74.68 & 72.87 \\
mSimCSE\textsubscript{MT\textsubscript{kr}} \\
\multicolumn{1}{l}{+ KorNLI} & 1.0M & \textbf{80.29} & 74.93 & \textbf{75.43} & \textbf{74.45} \\
\bottomrule
\end{tabular}
}
\quad
\subfloat[Both the premise and hypothesis.]{\label{tab:hypothesis}
\begin{tabular}{l@{}ccccc}
\toprule
& & English & Japanese & \multicolumn{2}{c}{Korean} \\
\multicolumn{1}{c}{Models} & Size & STS-B & \makecell{JGLUE-\\JSTS} & KorSTS & \makecell{KLUE-\\STS} \\
\midrule
\multicolumn{6}{c}{\textit{Unsupervised}} \\
XLM-R\textsubscript{base} & & 53.62 & 59.28 & 57.98 & 30.69 \\
mSimCSE\textsubscript{en} \\
\multicolumn{1}{l}{+ SNLI} & 1.0M & 75.03 & 72.69 & 73.71 & 58.75 \\
\multicolumn{1}{l}{+ NLI} & 2.0M & 75.83 & 73.12 & 74.18 & 61.75 \\
mSimCSE\textsubscript{MT\textsubscript{ja}} \\
\multicolumn{1}{l}{+ JSNLI} & 1.0M & 75.26 & 71.69 & 74.20 & 65.39 \\
\midrule
XLM-R\textsubscript{large} & & 43.54 & 54.19 & 49.26 & 20.13 \\
mSimCSE\textsubscript{en} \\
\multicolumn{1}{l}{+ SNLI} & 1.0M & 75.36 & 74.73 & 75.21 & 65.83 \\
\multicolumn{1}{l}{+ NLI} & 2.0M & 76.87 & 75.19 & 74.19 & 70.60 \\
mSimCSE\textsubscript{MT\textsubscript{ja}} \\
\multicolumn{1}{l}{+ JSNLI} & 1.0M & 79.40 & \textbf{75.55} & 75.21 & 73.97 \\
\bottomrule
\end{tabular}
}
\caption{
Ablation study on different data sizes, data combinations, and base models.
\textbf{Boldface} only highlights the best performance values over the two combinations of the premise alone and both the premise and hypothesis.
}\label{tab:ablation}
\end{table*}

\end{document}